\pdfoutput=1
\documentclass[11pt,letterpaper]{article}

\newcommand{\paperauthormain}{Chris Fournier}
\newcommand{\paperauthorsuperisor}{Diana Inkpen}
\newcommand{\papertitle}{Segmentation Similarity and Agreement}

\usepackage[pdfauthor={\paperauthormain},%
pdftitle={\papertitle},%
pdfborder=0 0 0
pdftex]{hyperref}

\hypersetup{ 
colorlinks,% 
citecolor=black,% 
filecolor=black,% 
linkcolor=black,% 
urlcolor=black 
}

\usepackage{naaclhlt2012}
\usepackage{times}
\usepackage{paralist}
\usepackage[caption=false]{subfig}
\usepackage{float}
\usepackage{latexsym}
\usepackage{amsmath}
\usepackage{amssymb}
\usepackage{amsfonts}
\usepackage{mathtools}
\usepackage{comment}
\usepackage{color}
\usepackage{tabularx}
\usepackage{subfig}
\usepackage[table, x11names, rgb]{xcolor}
\usepackage{tikz}
\usepackage{pgfplots}
\usetikzlibrary{decorations.pathreplacing,calc,positioning}

\usepackage{graphicx,eso-pic,xcolor}
\title{\papertitle}

\author{\paperauthormain \\
        University of Ottawa \\
 	    Ottawa, ON, Canada\\
	    {\tt cfour037@eecs.uottawa.ca}
	    \And
		\paperauthorsuperisor\\
  		University of Ottawa \\
 	    Ottawa, ON, Canada\\
	    {\tt diana@eecs.uottawa.ca}}

\date{}

\newcommand{\eg}{\textit{e.g.},~}
\newcommand{\ie}{\textit{i.e.},~}

\newcommand{\fmeasure}{{$\text{F}_{\beta}\text{-measure}$}}
\newcommand{\Pk}{$\text{P}_k$}
\newcommand{\Pu}{$\text{P}_\mu$}
\newcommand{\KazantsevaSzpakowicz}{Kazantseva~and~Szpakowicz}
\newcommand{\softwareurl}{\url{http://nlp.chrisfournier.ca/}}
\newcommand{\metricname}{segmentation similarity} % our
\newcommand{\metricacronym}{S} % 
\newcommand{\errorunit}{PB} % 
\newcommand{\errorunits}{PBs} % 
\newcommand{\sizeunit}{unit} % 
\newcommand{\sizeunits}{units} % 
\newcommand{\oneminuswd}{$1$$-$$WD$} % 

\tikzset{
  region/.style={decorate,decoration={brace,amplitude=0.35em}},
  region_beneath/.style={decorate,decoration={brace,mirror,amplitude=0.35em,raise=1.3em}}
}

\colorlet{block_norm}{gray!15}
\colorlet{block_miss}{gray!70}
\colorlet{block_white}{white}
\colorlet{label_colour}{black}

\newcommand{\spacer}{\vspace{-0.40em}}
\newcommand{\spacersmall}{\vspace{-0.20em}}

\tikzstyle{header}=[draw, rectangle, minimum width=1.25em, minimum
height=1.25em, anchor=south west]
\tikzstyle{bound}=[draw, rectangle, minimum width=1.25em, minimum
height=1.25em, anchor=south west]
\tikzstyle{header_bound}=[draw=none, minimum width=1.75em, minimum
height=1.25em, anchor=south west, xshift=0.65em]
\tikzstyle{bound}=[draw=none, minimum width=1.75em, minimum
height=1.25em, anchor=south west, xshift=-0.5em]
\tikzstyle{bound_empty}=[draw=none, minimum width=1.75em, minimum
height=1.25em, anchor=south west, xshift=-0.5em]
\tikzstyle{block1}=[draw, rectangle, minimum width=1.25em, minimum
height=1.25em, fill=block_norm,anchor=south west]
\tikzstyle{block1mis}=[draw, rectangle, minimum width=1.25em, minimum
height=1.25em, fill=block_miss,anchor=south west]
\tikzstyle{block2}=[draw, rectangle, minimum width=2.50em, minimum
height=1.25em, fill=block_norm,anchor=south west]
\tikzstyle{block2mis}=[draw, rectangle, minimum width=2.50em, minimum
height=1.25em, fill=block_miss,anchor=south west]
\tikzstyle{block3}=[draw, rectangle, minimum width=3.75em, minimum
height=1.25em, fill=block_norm,anchor=south west]
\tikzstyle{block3mis}=[draw, rectangle, minimum width=3.75em, minimum
height=1.25em, fill=block_miss,anchor=south west]
\tikzstyle{block4}=[draw, rectangle, minimum width=5.00em, minimum
height=1.25em, fill=block_norm,anchor=south west]
\tikzstyle{block4mis}=[draw, rectangle, minimum width=5.00em, minimum
height=1.25em, fill=block_miss,anchor=south west]
\tikzstyle{block5}=[draw, rectangle, minimum width=6.25em, minimum
height=1.25em, fill=block_norm,anchor=south west]
\tikzstyle{block6}=[draw, rectangle, minimum width=7.50em, minimum
height=1.25em, fill=block_norm,anchor=south west]
\tikzstyle{block6mis}=[draw, rectangle, minimum width=7.50em, minimum
height=1.25em, fill=block_miss,anchor=south west]
\tikzstyle{block7}=[draw, rectangle, minimum width=8.75em, minimum
height=1.25em, fill=block_norm,anchor=south west]
\tikzstyle{block8}=[draw, rectangle, minimum width=10.00em, minimum
height=1.25em, fill=block_norm,anchor=south west]
\tikzstyle{block9}=[draw, rectangle, minimum width=11.25em, minimum
height=1.25em, fill=block_norm,anchor=south west]
\tikzstyle{block10}=[draw, rectangle, minimum width=12.50em, minimum
height=1.25em, fill=block_norm,anchor=south west]
\tikzstyle{block11}=[draw, rectangle, minimum width=13.75em, minimum
height=1.25em, fill=block_norm,anchor=south west]
\tikzstyle{block12}=[draw, rectangle, minimum width=15.00em, minimum
height=1.25em, fill=block_norm,anchor=south west]
\tikzstyle{block13}=[draw, rectangle, minimum width=16.25em, minimum
height=1.25em, fill=block_norm,anchor=south west]
\tikzstyle{block14}=[draw, rectangle, minimum width=17.50em, minimum
height=1.25em, fill=gray!20,anchor=south west]
\tikzstyle{block15}=[draw, rectangle, minimum width=18.75em, minimum
height=1.25em, fill=gray!20,anchor=south west]
\tikzstyle{block16}=[draw, rectangle, minimum width=20.00em, minimum
height=1.25em, fill=gray!20,anchor=south west]
\tikzstyle{label}=[minimum width=1.25em, minimum height=1.25em, anchor=south
west, color=black!60]
\tikzstyle{label1}=[minimum width=1.30em, minimum height=1.25em, anchor=south
west, text centered, color=label_colour]
\tikzstyle{label2}=[minimum width=3em, minimum height=1.25em, anchor=south
west, text centered, color=label_colour]
\tikzstyle{label6}=[minimum width=7.80em, minimum height=1.25em, anchor=south
west, text centered, color=label_colour]
\tikzstyle{becomes}=[minimum width=1.25em, minimum height=2.5em, anchor=south
west]
\tikzstyle{edit1}=[minimum width=1.30em, minimum height=1.25em, anchor=south
west, text centered, color=black]

\begin{document}
\maketitle
\begin{abstract}
We propose a new segmentation evaluation metric, called \emph{\metricname}
(\metricacronym), that quantifies the similarity between two segmentations as
the proportion of boundaries that are not transformed when comparing them using
edit distance, essentially using edit distance as a penalty function and scaling
penalties by segmentation size. We propose several adapted inter-annotator
agreement coefficients which use \metricacronym{} that are suitable for
segmentation.  We show that \metricacronym{} is configurable enough to
suit a wide variety of segmentation evaluations, and is an improvement upon the
state of the art. We also propose using inter-annotator agreement
coefficients to evaluate automatic segmenters in terms of human performance.
\end{abstract}

\section{Introduction}
% Introduce segmentation, automatic segmentation, and the need for human-
% competititve perfomance.
Segmentation is the task of splitting up an item, such as a document, into a
sequence of segments by placing boundaries within.  The purpose of segmenting
can vary greatly, but one common objective is to denote shifts in the topic of a
text, where multiple boundary types can also be present (\eg major versus minor
topic shifts). Human-competitive automatic segmentation methods can help a wide
range of computational linguistic tasks which depend upon the identification of
segment boundaries in text.

% Identify the need for comparison to multiple coders
To evaluate automatic segmentation methods, a method of comparing an automatic
segmenter's performance against the segmentations produced by human judges
(coders) is required.  Current methods of performing this comparison designate
only one coder's segmentation as a reference to compare against.  A single ``true''
reference segmentation from a coder should not be trusted, given that
inter-annotator agreement is often reported to be rather poor \cite[p.
54]{Hearst:1997}.  Additionally, to ensure that an automatic segmenter does not
over-fit to the preference and bias of one particular coder, an automatic
segmenter should be compared directly against multiple coders.

% Identify the need for a replacement
The state of the art segmentation evaluation metrics (\Pk{} and WindowDiff)
slide a window across a designated reference and hypothesis segmentation, and
count the number of windows where the number of boundaries differ.  Window-based
methods suffer from a variety of problems, including:
\begin{inparaenum}[i\upshape)]
\item unequal penalization of error types;
\item an arbitrarily defined window size parameter (whose choice greatly
affects outcomes);
\item lack of clear intuition;
\item inapplicability to multiply-coded corpora; and
\item reliance upon a ``true'' reference segmentation.
\end{inparaenum}

% Propose similarity and inter-annotator agreement
In this paper, we propose a new method of comparing two segmentations,
called \emph{\metricname} \footnote{A software implementation of segmentation
similarity (\metricacronym{}) is available at \softwareurl{}} (\metricacronym),
that:
\begin{inparaenum}[i\upshape)]
\item equally penalizes all error types (unless explicitly configured
otherwise);
\item appropriately responds to scenarios tested;
\item defines no arbitrary parameters;
\item is intuitive; and
\item is adapted for use in a variety of popular inter-annotator agreement
coefficients to handle multiply-coded corpora; and
\item does not rely upon a ``true'' reference segmentation (it is symmetric).
\end{inparaenum} Capitalizing on the
adapted inter-annotator agreement coefficients, the relative difficulty that
human segmenters have with various segmentation tasks can now be quantified.
We also propose that these coefficients can be used to evaluate and compare
automatic segmentation methods in terms of human agreement.

% Outline
This paper is organized as follows.  In Section~\ref{section:related.work},
we review segmentation evaluation and inter-annotator agreement. In
Section~\ref{section:similarity}, we present \metricacronym{} and
inter-annotator agreement coefficient adaptations.  In
Section~\ref{section:experiments}, we evaluate \metricacronym{} and
WindowDiff in various scenarios and simulations, and upon a multiply-coded
corpus.

\section{Related Work}\label{section:related.work}

\subsection{Segmentation Evaluation}
% Precision, recall, f-measure
Precision, recall, and their mean (\fmeasure{}) have been previously applied to
segmentation evaluation. Precision is the proportion of boundaries chosen that
agree with a reference segmentation, and recall is the proportion of boundaries
chosen that agree with a reference segmentation out of all boundaries in the
reference and hypothesis \cite[p. 3]{PevznerHearst:2002}. For segmentation,
these metrics are unsuitable because they penalize near-misses of boundaries as
full-misses, causing them to drastically overestimate the error.  Near-misses
are prevalent in segmentation and can account for a large proportion of the errors
produced by a coder, and as inter-annotator agreement often shows, they do not
reflect coder error, but the difficulty of the task.

% Pk
\Pk~\cite[pp. 198--200]{BeefermanBerger:1999}\footnote{\Pk~is a modification of
\Pu~\cite[p. 43]{BeefermanEtAl:1997}. Other modifications such as TDT
$C_{seg}$~\cite[pp. 5--6]{Doddington:1998} have been proposed, but \Pk~has seen
greater usage.}
is a window-based metric which attempts to solve the harsh near-miss
penalization of precision, recall, and \fmeasure.  In \Pk, a window of size $k$,
where $k$ is defined as half of the mean reference segment size, is slid across
the text to compute penalties.  A penalty of 1 is assigned for each window whose
boundaries are detected to be in different segments of the reference and
hypothesis segmentations, and this count is normalized by the number of windows.

Pevzner~and~Hearst~\shortcite[pp. 5--10]{PevznerHearst:2002} highlighted a
number of issues with \Pk, specifically that:
\begin{inparaenum}[i\upshape)]
\item False negatives (FNs) are penalized more than false positives (FPs);
\item It does not penalize FPs that fall within $k$ units of a reference
boundary;
\item Its sensitivity to variations in segment size can cause it to linearly
decrease the penalty for FPs if the size of any segments fall below $k$; and
\item Near-miss errors are too harshly penalized.
\end{inparaenum}

% WindowDiff
To attempt to mitigate the shortcomings of \Pk,
Pevzner~and~Hearst~\shortcite[p. 10]{PevznerHearst:2002} proposed a modified
metric which changed how penalties were counted, named \emph{WindowDiff} ($WD$).
A window of size $k$ is still slid across the text, but now penalties are
attributed to windows where the number of boundaries in each segmentation
differs (see Equation~\ref{eqn:windowdiff}, where $\text{b}(R_{ij})$ and
$\text{b}(H_{ij})$ represents the number of boundaries within the segments in a
window of size $k$ from position $i$ to $j$, and $N$ the number of sentences
plus one), with the same normalization.

\begin{footnotesize}
\spacer
\spacersmall
\begin{equation}\label{eqn:windowdiff}
\text{WD}(R,H) = \frac{1}{N-k}
\sum^{N-k}_{i=1,j=i+k}(|\text{b}(R_{ij})-\text{b}(H_{ij})| > 0)
\end{equation}
\spacersmall
\spacer
\end{footnotesize}

WindowDiff is able to reduce, but not eliminate, sensitivity to segment size,
gives more equal weights to both FPs and FNs (FNs are, in effect, penalized
less\footnote{Georgescul~et~al.~\shortcite[p. 48]{GeorgesculEtAl:2006}
note that both FPs and FNs are weighted by $^1/_{N-k}$, and although there
are ``equiprobable possibilities to have a [FP] in an interval of $k$ units'',
``the total number of equiprobable possibilities to have a [FN] in an interval
of $k$ units is smaller than ($N-k$)'', making the interpretation of a full miss
as a FN less probable than as a FP.}), and is
able to catch mistakes in both small and large segments.  It is not
without issues though; Lamprier~et~al.~\shortcite{LamprierEtAl:2007}
demonstrated that WindowDiff penalizes errors less at the beginning and end of a
segmentation (this is corrected by padding the segmentation at each end by size
$k$). Additionally, variations in the window size $k$ lead to difficulties in interpreting and
comparing WindowDiff's values, and the intuition of the method remains vague.

Franz~et~al.~\shortcite{FranzEtAl:2007} proposed measuring performance in terms
of the number of words that are FNs and FPs, normalized by the number of word
positions present (see Equation~\ref{eqn:r}).

\begin{footnotesize}
\spacer
\spacersmall
\begin{equation}\label{eqn:r}
R_{FN} = \frac{1}{N}\sum_w{FN(w)},\quad R_{FP} = \frac{1}{N}\sum_w{FP(w)}
\end{equation}
\spacersmall
\spacer
\end{footnotesize}

$R_{FN}$ and $R_{FP}$ have the advantage that they take into account the
severity of an error in terms of segment size, allowing them to reflect the
effects of erroneously missing, or added, words in a segment better than window
based metrics.  Unfortunately, $R_{FN}$ and $R_{FP}$ suffer from the same flaw
as precision, recall, and \fmeasure~in that they do not account for near misses.

\subsection{Inter-Annotator Agreement}
The need to ascertain the agreement and reliability between coders for
segmentation was recognized by
Passonneau~and~Litman~\shortcite{PassonneauLitman:1993}, who adapted the
percentage agreement metric by Gale~et~al.~\shortcite[p. 254]{GaleEtAl:1992}
for usage in segmentation. This percentage agreement metric \cite[p.
150]{PassonneauLitman:1993} is the ratio of the total observed agreement of a
coder with the majority opinion for each boundary over the total possible
agreements.  This measure failed to take into account chance agreement, or to
less harshly penalize near-misses.

Hearst~\shortcite{Hearst:1997} collected segmentations from 7 coders while
developing the automatic segmenter TextTiling, and reported mean
$\kappa$~\cite{SiegelCastellan:1988} values for coders and automatic segmenters
\cite[p. 56]{Hearst:1997}.  Pairwise mean $\kappa$ scores were calculated by
comparing a coder's segmentation against a reference segmentation formulated by
the majority opinion strategy used in Passonneau~and~Litman~\shortcite[p.
150]{PassonneauLitman:1993} \cite[pp. 53--54]{Hearst:1997}.  Although mean
$\kappa$ scores attempt to take into account chance agreement, near misses are
still unaccounted for, and use of Siegel and Castellan's
\shortcite{SiegelCastellan:1988} $\kappa$ has declined in favour of other
coefficients~\cite[pp. 555--556]{ArtsteinPoesio:2008}.

Artstein and Poesio~\shortcite{ArtsteinPoesio:2008} briefly touch upon
recommendations for coefficients for segmentation evaluation, and though they do
not propose a measure, they do conjecture that a modification of a weighted form
of $\alpha$~\cite{Krippendorff:1980,Krippendorff:2004} using unification and
WindowDiff may suffice ~\cite[pp. 580--582]{ArtsteinPoesio:2008}.

\section{Segmentation Similarity}\label{section:similarity}
For discussing segmentation, a segment's size (or mass) is measured in units,
the error is quantified in potential boundaries ($\text{\errorunits{}}$), and
we have adopted a modified form of the notation used by Artstein and
Poesio~\shortcite{ArtsteinPoesio:2008}, where the set of:
\begin{itemize}
  \small
  \setlength{\itemsep}{0.0em}%
  \setlength{\parskip}{0.0em}%
  \item \emph{Items} is $\{i|i \in I\}$
  with cardinality \textbf{i};
  \item \emph{Categories} is $\{k|k \in K\}$ with cardinality \textbf{k};
  \item \emph{Coders} is $\{c|c \in C\}$ with cardinality
  \textbf{c};
  \item \emph{Segmentations} of an item $i$ by a coder $c$ is $\{s|s \in S\}$,
  where when $s_{ic}$  is specified with only one subscript, it denotes $s_{c}$,
  for all relevant items ($i$); and
  \item \emph{Types} of segmentation boundaries is $\{t|t \in T\}$ with
  cardinality \textbf{t}.
\end{itemize}

\subsection{Sources of Dissimilarity}\label{section:dissimilarity}
Linear segmentation has three main types of errors:
\begin{enumerate}
  \small
  \setlength{\itemsep}{0.0em}%
  \setlength{\parskip}{0.0em}%
  \item $s_1$ contains a boundary that is off by $n$ \errorunits{} in $s_2$;
  \item $s_1$ contains a boundary that $s_2$ does not; or
  \item $s_2$ contains a boundary that $s_1$ does not.
\end{enumerate}

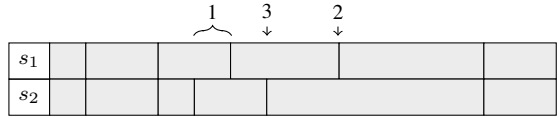
\begin{figure}[h]
  \centering
  \spacersmall
  \spacer
\begin{tikzpicture}[node distance=0cm, outer sep=0pt, font=\scriptsize,
scale=0.7]
\node[header] (a) at (1,8) 		   {\scriptsize{$s_1$}};
\node[header] (b)  [below = of a]  {\scriptsize{$s_2$}};
\node[block1] (r1) [right = of a]  {};
\node[block2] (r2) [right = of r1] {};
\node[block2] (r3) [right = of r2] {};
\node[block3] (r4) [right = of r3] {};
\node[block4] (r5) [right = of r4] {};
\node[block2] (r6) [right = of r5] {};
\node[block1] (h1) [right = of b]  {};
\node[block2] (h2) [right = of h1] {};
\node[block1] (h3) [right = of h2] {};
\node[block2] (h4) [right = of h3] {};
\node[block6] (h5) [right = of h4] {};
\node[block2] (h6) [right = of h5] {};
\draw[region] let
    \p1=(r3.west), \p2=(r4.west) in
    ($(\x1+1.75em,\y1+1.25em)$) -- ($(\x2+0.05em,\y2+1.25em)$) node[above,
    outer sep=0.25em, midway] {1};
\draw[->] let
    \p1=(r3.east), \p2=(r3.east) in
    ($(\x1+1.8em,\y1+1.7em)$) -- ($(\x2+1.8em,\y2+1.25em)$) node[above,
    outer sep=0.1em, midway] {3};
\draw[->] let
    \p1=(r3.east), \p2=(r3.east) in
    ($(\x1+5.3em,\y1+1.7em)$) -- ($(\x2+5.3em,\y2+1.25em)$) node[above,
    outer sep=0.1em, midway] {2};
\end{tikzpicture}
  \spacer
  \spacer
  \caption{Types of segmentations errors}
  \label{fig:segment.mass.unlabelled}
  \spacer
  \spacer
\end{figure}

These types of errors can be seen in Figure~\ref{fig:segment.mass.unlabelled},
and are conceptualized as a pairwise \emph{transposition} of a boundary for
error 1, and the insertion or deletion (depending upon your perspective) of a
boundary for errors 2 and 3.  Since we do not designate
either segmentation as a reference or hypothesis, we refer to
insertions and deletions both as \emph{substitutions}.

It is important to not penalize near misses as full misses in many segmentation
tasks because coders often agree upon the existence of a boundary, but disagree
upon its exact location.  In the previous scenario, assigning a full miss would
mean that even a boundary loosely agreed-upon, as in
Figure~\ref{fig:segment.mass.unlabelled}, error 1, would be regarded
as completely disagreed-upon.

\subsection{Edit Distance}\label{section:editdistance}
% Detail my edit distance metric
In \metricacronym{}, concepts from Damereau-Levenshtein edit
distance~\cite{Damerau:1964,Levenshtein:1966} are applied to model segmentation
edit distance as two operations: substitutions and 
transpositions.\footnote{Beeferman~et~al.~\shortcite[p. 42]{BeefermanEtAl:1997}
briefly mention using an edit distance without transpositions, but discard it in favour of \Pu{}.}
These two operations represent full misses
and near misses, respectively.  Using these two operations, a new
globally-optimal minimum edit distance is applied to a pair of sequences of sets
of boundaries to model the sources of dissimilarity identified
earlier.\footnote{For multiple boundaries, an \emph{add/del} operation is added,
and transpositions are considered only within boundary types.}

Near misses that are remedied by transposition are penalized as $b$ \errorunits{}
of error (where $b$ is the number of boundaries transposed), as opposed to the
$2b$ \errorunits{} of errors by which they would be penalized if they were
considered to be two separate substitution operations. Transpositions can also
be considered over $n>2$ \errorunits{} ($n$-wise transpositions).  This is useful
if, for a specific task, near misses of up to $n$ \errorunits{} are not
to be penalized as full misses (default $n=2$).

The error represented by the two operations can also be scaled (\ie weighted)
from 1 \errorunit{} each to a fraction.  The distance over which an $n$-wise
transposition occurred can also be used in conjunction with the scalar operation
weighting so that a transposition is weighted using the function in
Equation~\ref{eqn:transposition.penalty}.

\begin{footnotesize}
\spacer
\spacersmall
\begin{alignat}{1}\label{eqn:transposition.penalty}
\text{te}(n,b)=b-(^1/_b)^{n-2} \quad \text{where } n \geq 2 \text{ and } b > 0
\end{alignat}
\vspace{-1.5em}
\end{footnotesize}

This transposition error function was
chosen so that, in an $n$-wise transposition where $n=2$ \errorunits{} and the
number of boundaries transposed $b=2$, the penalty would be 1 \errorunit{}, and
the maximum penalty as $\lim_{n \to \infty} \text{te}(n)$ would be $b$
\errorunits{}, or in this case 2 \errorunits{} (demonstrated later in
Figure~\ref{fig:increasing.transp.dist}).

\subsection{Method}\label{section:method}
In \metricacronym{}, we conceptualize the entire segmentation, and
individual segments, as having mass (\ie unit magnitude/length), and quantify
similarity between two segmentations as the proportion of boundaries that are
not transformed when comparing segmentations using edit distance, essentially
using edit distance as a penalty function and scaling penalties by segmentation
size.  \metricacronym{} is a symmetric function that quantifies the similarity
between two segmentations as a percentage, and applies to any granularity or
segmentation unit (\eg paragraphs, sentences, clauses, etc.).

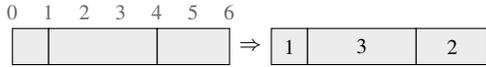
\begin{figure}[h]
  \centering
\spacersmall
\begin{tikzpicture}[node distance=0cm, outer sep=0pt, font=\scriptsize,
scale=0.7]
\node[block1] (r1) at (1,8) 		{};
\node[block3] (r2) [right = of r1]  {};
\node[block2] (r3) [right = of r2]  {};
\node[label]  (l1) at (0.65,8.65) 	{0};
\node[label]  (l2) [right = of l1]  {1};
\node[label]  (l3) [right = of l2]  {2};
\node[label]  (l4) [right = of l3]  {3};
\node[label]  (l5) [right = of l4]  {4};
\node[label]  (l6) [right = of l5]  {5};
\node[label]  (l7) [right = of l6]  {6};
\node[becomes](ref)[right = of r3]  {$\Rightarrow$};
\node[block1] (c1) [right = of ref] {1};
\node[block3] (c2) [right = of c1]  {3};
\node[block2] (c3) [right = of c2]  {2};
\end{tikzpicture}
\spacer
\spacer
  \caption{Annotation of segmentation mass}
  \label{fig:segment.mass}
\spacersmall
\end{figure}

Consider a somewhat contrived example containing--for simplicity and
brevity--only one boundary type ($\textbf{t}=1$). First, a segmentation must be
converted into a sequence of segment mass values (see
Figure~\ref{fig:segment.mass}).

Then, a pair of segmentations are converted into parallel sequences of boundary
sets, where each set contains the types of boundaries present at that potential
boundary location (if there is no boundary present, then the set is empty), as
in Figure~\ref{fig:sequences}.

% [1,2,2,3,3,1,2][1,2,1,2,6,2]
\begin{figure}[h]
  \centering
\begin{tikzpicture}[node distance=0cm, outer sep=0pt, font=\small,
scale=0.7]
\node[header] (a) at (1,8) 		   {\scriptsize{$s_1$}};
\node[header] (b)  [below = of a]  {\scriptsize{$s_2$}};
\node[block1] (r1) [right = of a]  {1};
\node[block2] (r2) [right = of r1] {2};
\node[block2] (r3) [right = of r2] {2};
\node[block3] (r4) [right = of r3] {3};
\node[block3] (r5) [right = of r4] {3};
\node[block1] (r6) [right = of r5] {1};
\node[block2] (r7) [right = of r6] {2};
\node[header_bound] (b1) [above = of r1]  {\tiny{$\{1\}$}};
\node[bound_empty]	(b2) [right = of b1]  {\tiny{$\{ \}$}};
\node[bound]		(b3) [right = of b2]  {\tiny{$\{1\}$}};
\node[bound_empty]	(b4) [right = of b3]  {\tiny{$\{ \}$}};
\node[bound]		(b5) [right = of b4]  {\tiny{$\{1\}$}};
\node[bound_empty]	(b6) [right = of b5]  {\tiny{$\{ \}$}};
\node[bound_empty]	(b7) [right = of b6]  {\tiny{$\{ \}$}};
\node[bound]		(b8) [right = of b7]  {\tiny{$\{1\}$}};
\node[bound_empty]	(b9) [right = of b8]  {\tiny{$\{ \}$}};
\node[bound_empty]	(b10)[right = of b9]  {\tiny{$\{ \}$}};
\node[bound] 		(b11)[right = of b10] {\tiny{$\{1\}$}};
\node[bound] 		(b12)[right = of b11] {\tiny{$\{1\}$}};
\node[bound_empty]	(b13)[right = of b12] {\tiny{$\{ \}$}};
\node[block1] (h1) [right = of b]  {1};
\node[block2] (h2) [right = of h1] {2};
\node[block1] (h3) [right = of h2] {1};
\node[block2] (h4) [right = of h3] {2};
\node[block6] (h5) [right = of h4] {6};
\node[block2] (h6) [right = of h5] {2};
\node[header_bound] (c1) [below = of h1]  {\tiny{$\{1\}$}};
\node[bound_empty]	(c2) [right = of c1]  {\tiny{$\{ \}$}};
\node[bound]		(c3) [right = of c2]  {\tiny{$\{1\}$}};
\node[bound]		(c4) [right = of c3]  {\tiny{$\{1\}$}};
\node[bound_empty]	(c5) [right = of c4]  {\tiny{$\{ \}$}};
\node[bound]		(c6) [right = of c5]  {\tiny{$\{1\}$}};
\node[bound_empty]	(c7) [right = of c6]  {\tiny{$\{ \}$}};
\node[bound_empty]	(c8) [right = of c7]  {\tiny{$\{ \}$}};
\node[bound_empty]	(c9) [right = of c8]  {\tiny{$\{ \}$}};
\node[bound_empty]	(c10)[right = of c9]  {\tiny{$\{ \}$}};
\node[bound_empty] 	(c11)[right = of c10] {\tiny{$\{ \}$}};
\node[bound] 		(c12)[right = of c11] {\tiny{$\{1\}$}};
\node[bound_empty]	(c13)[right = of c12] {\tiny{$\{ \}$}};
\end{tikzpicture}
\spacer
\spacer
  \caption{Segmentations annotated with mass and their corresponding boundary set sequences}
  \label{fig:sequences}
\end{figure}

The edit distance is calculated by first identifying all potential substitution
operations that could occur (in this case 5).  A search for all potential
$n$-wise transpositions that can be made over $n$ adjacent sets between the
sequences is then performed, searching from the beginning of the sequence to the
end, keeping only those transpositions which do not overlap and which result in
transposing the most boundaries between the sequences (to minimize the edit
distance).  In this case, we have only one non-overlapping 2-wise transposition.
 We then subtract the number of boundaries involved in transpositions between
the sequences (2 boundaries) from the number of substitutions, giving us an edit
distance of 4 \errorunits{}: 1 transposition \errorunit{} and 3 substitution
\errorunits{}.

% [1,2,2,3,3,1,2][1,2,1,2,6,2]
\begin{figure}[h]
  \centering
\spacer
\spacer
\begin{tikzpicture}[node distance=0cm, outer sep=0pt, font=\small,
scale=0.7]
\node[header] (a) at (1,8) 		   {\scriptsize{$s_1$}};
\node[header] (b)  [below = of a]  {\scriptsize{$s_2$}};
\node[block1] (r1) [right = of a]  {1};
\node[block2] (r2) [right = of r1] {2};
\node[block2] (r3) [right = of r2] {2};
\node[block3] (r4) [right = of r3] {3};
\node[block3] (r5) [right = of r4] {3};
\node[block1] (r6) [right = of r5] {1};
\node[block2] (r7) [right = of r6] {2};
\node[header_bound] (b1) [above = of r1]  {\tiny{$\{1\}$}};
\node[bound_empty]	(b2) [right = of b1]  {\tiny{$\{ \}$}};
\node[bound]		(b3) [right = of b2]  {\tiny{$\{1\}$}};
\node[bound_empty]	(b4) [right = of b3]  {\tiny{$\{ \}$}};
\node[bound]		(b5) [right = of b4]  {\tiny{$\{1\}$}};
\node[bound_empty]	(b6) [right = of b5]  {\tiny{$\{ \}$}};
\node[bound_empty]	(b7) [right = of b6]  {\tiny{$\{ \}$}};
\node[bound]		(b8) [right = of b7]  {\tiny{$\{1\}$}};
\node[bound_empty]	(b9) [right = of b8]  {\tiny{$\{ \}$}};
\node[bound_empty]	(b10)[right = of b9]  {\tiny{$\{ \}$}};
\node[bound] 		(b11)[right = of b10] {\tiny{$\{1\}$}};
\node[bound] 		(b12)[right = of b11] {\tiny{$\{1\}$}};
\node[bound_empty]	(b13)[right = of b12] {\tiny{$\{ \}$}};
\node[block1] (h1) [right = of b]  {1};
\node[block2] (h2) [right = of h1] {2};
\node[block1] (h3) [right = of h2] {1};
\node[block2] (h4) [right = of h3] {2};
\node[block6] (h5) [right = of h4] {6};
\node[block2] (h6) [right = of h5] {2};
\node[header_bound] (c1) [below = of h1]  {\tiny{$\{1\}$}};
\node[bound_empty]	(c2) [right = of c1]  {\tiny{$\{ \}$}};
\node[bound]		(c3) [right = of c2]  {\tiny{$\{1\}$}};
\node[bound]		(c4) [right = of c3]  {\tiny{$\{1\}$}};
\node[bound_empty]	(c5) [right = of c4]  {\tiny{$\{ \}$}};
\node[bound]		(c6) [right = of c5]  {\tiny{$\{1\}$}};
\node[bound_empty]	(c7) [right = of c6]  {\tiny{$\{ \}$}};
\node[bound_empty]	(c8) [right = of c7]  {\tiny{$\{ \}$}};
\node[bound_empty]	(c9) [right = of c8]  {\tiny{$\{ \}$}};
\node[bound_empty]	(c10)[right = of c9]  {\tiny{$\{ \}$}};
\node[bound_empty] 	(c11)[right = of c10] {\tiny{$\{ \}$}};
\node[bound] 		(c12)[right = of c11] {\tiny{$\{1\}$}};
\node[bound_empty]	(c13)[right = of c12] {\tiny{$\{ \}$}};
\draw[region_beneath] let
    \p1=(c4.west), \p2=(c5.east) in
    ($(\x1+0.75em,\y1+1.05em)$) -- ($(\x2-0.75em,\y2+1.05em)$) node[below, midway, yshift=-1.5em] 
    {Transposition};
\draw[->] let
    \p1=(b6.east), \p2=(b6.east) in
    ($(\x1-1.3em,\y1+1.4em)$) -- ($(\x2-1.3em,\y2+0.75em)$) node[above, yshift=0.1em, midway] {Sub.};
\draw[->] let
    \p1=(b8.east), \p2=(b8.east) in
    ($(\x1-1.3em,\y1+1.4em)$) -- ($(\x2-1.3em,\y2+0.75em)$) node[above, yshift=0.1em, midway] {Sub.};
\draw[->] let
    \p1=(b11.east), \p2=(b11.east) in
    ($(\x1-1.3em,\y1+1.4em)$) -- ($(\x2-1.3em,\y2+0.75em)$) node[above, yshift=0.1em, midway] {Sub.};
\end{tikzpicture}
\spacer
\spacer
  \caption{Edit operations performed on boundary sets}
  \label{fig:sequences.labelled}
\spacer
\spacersmall
\end{figure}

Edit distance, and especially the number of operations of each type performed,
is  useful in identifying the number of full and near misses that
have occurred--which indicates whether one's choice of transposition window
size $n$ is either too generous or too harsh. Edit distance as a penalty
does not incorporate information on the severity of an error with respect to
the size of a segment, and is not an easily comparable value without some
form of normalization.  To account for these issues, we define S so that
boundary edit distance is used to subtract penalties for each edit operation that occurs,
from the number of potential boundaries in a segmentation, normalizing this
value by the total number of potential boundaries in a segmentation.

\begin{figure}[h]
\spacer
\begin{footnotesize}
\begin{alignat}{1}\label{eqn:similarity}
\hspace{-0.5em}
\text{S}(s_{i1},s_{i2}) = \frac
{\textbf{t} \cdot \text{mass}(i) - \textbf{t} - \text{d}(s_{i1},s_{i2},T)}
{\textbf{t} \cdot \text{mass}(i) - \textbf{t}}
\end{alignat}
\end{footnotesize}
\spacer
\spacer
\spacer
\spacersmall
\end{figure}

\metricacronym{}, as shown in Equation~\ref{eqn:similarity}, scales the mass of
the item by the cardinality of the set of boundary types ($\textbf{t}$) because
the edit distance function $\text{d}(s_{i1},s_{i1},T)$ will return a value of
$[0,\textbf{t} \cdot \text{mass}(i)]$ \errorunits{}, where $\textbf{t} \in
\mathbb{Z}^+$--while subtracting the edit distance and
$\textbf{t}$.\footnote{The number of potential boundaries in a segmentation
$s_i$ with $\textbf{t}$ boundary types is $\textbf{t} \cdot \text{mass}(i) -
\textbf{t}$.} The numerator is normalized by the total number of potential
boundaries per boundary type. This results in a function with a range of
$[0,1]$. It returns 0 when one segmentation contains no boundaries, and the
other contains the maximum number of possible boundaries. It returns 1 when both
segmentations are identical.

Using the default configuration of this equation, \metricacronym{} $=
{^{9}/_{13}} = 0.6923$, a very low similarity, which WindowDiff also agrees upon
(\oneminuswd{} $ = 0.6154$).  The edit-distance function
$\text{d}(s_{i1},s_{i1},T)$ can also be assigned values of the range $[0,1]$ as
scalar weights ($\text{w}_{sub}$, $\text{w}_{trp}$) to reduce the penalty
attributed to particular edit operations, and configured to use a transposition
error function (Equation~\ref{eqn:transposition.penalty}, used by default).

\subsection{Evaluating Automatic Segmenters}
Coders often disagree in segmentation tasks \cite[p. 56]{Hearst:1997}, making
it improbable that a single, correct, reference segmentation could be identified
from human codings. This improbability is the result of individual coders
adopting slightly different segmentation strategies (\ie different granularity).
In light of this, we propose that the best available evaluation strategy
for automatic segmentation methods is to compare performance against multiple
coders directly, so that performance can be quantified relative to human
reliability and agreement.

To evaluate whether an automatic segmenter performs on par with human
performance, inter-annotator agreement can be calculated with and without the
inclusion of an automatic segmenter, where an observed drop in the coefficients
would signify that the automatic segmenter does not perform as reliably as the
group of human coders.\footnote{Similar to how human competitiveness is
ascertained by Medelyan~et~al.~\shortcite[pp.
1324--1325]{MedelyanEtAl:2009} and Medelyan~\shortcite[pp.
143--145]{Medelyan:2009} by comparing drops in inter-indexer consistency.} This
can be performed independently for multiple automatic segmenters to compare them
to each other--assuming that the coefficients model chance agreement
appropriately--because agreement is calculated (and quantifies reliability) over
all segmentations.

\subsection{Inter-Annotator Agreement}\label{section:agreement}
% Present case for agreement, and not pair-wise 
Similarity alone is not a sufficiently insightful measure of reliability, or
agreement, between coders. Chance agreement occurs in segmentation when coders
operating at slightly different granularities agree due to their codings, and
not their own innate segmentation heuristics.
Inter-annotator agreement coefficients have been developed that assume a variety
of prior distributions to characterize chance agreement, and to attempt to offer
a way to identify whether agreement is primarily due to chance, or not, and to
quantify reliability.

Artstein and Poesio~\shortcite{ArtsteinPoesio:2008} note that most of a coder's
judgements are non-boundaries. The class imbalance caused by segmentations often
containing few boundaries, paired with no handling of near misses, causes most
inter-annotator agreement coefficients to drastically underestimate agreement on
segmentations.  To allow for agreement coefficients to account for near misses,
we have adapted \metricacronym{} for use with Cohen's~$\kappa$,
Scott's~$\pi$,  Fleiss's~multi-$\pi$ ($\pi^*$), and Fleiss's~multi-$\kappa$
($\kappa^*$), which are all coefficients that range from $[\textstyle^{A_e}/_{1
- A_e},1]$, where 0 indicates chance agreement, and 1 perfect agreement. All four
coefficients have the general form:

\begin{footnotesize}
\spacersmall
\begin{equation}
\kappa,\pi,\kappa^*,\text{ and }\pi^* = \frac{\text{A}_a-\text{A}_e}{1 -
\text{A}_e}
\end{equation}
\spacersmall
\spacer
\spacer
\end{footnotesize}

For each agreement coefficient, the set of categories is defined as solely the
presence of a boundary ($K = \{\text{seg}_t | t \in T\}$), per boundary type
($t$). This category choice is similar to those chosen by Hearst~\shortcite[p.
53]{Hearst:1997}, who computed chance agreement in terms of the probability that
coders would say that a segment boundary exists ($\text{seg}_t$),
and the probability that they would not ($\text{unseg}_t$).  We have
chosen to model chance agreement only in terms of the presence of a boundary,
and not the absence, because coders have only two choices when
segmenting: to place a boundary, or not.  Coders do not place non-boundaries.  If they do not
make a choice, then the default choice is used: no boundary. This
default option makes it impossible to determine whether a segmenter
is making a choice by not placing a boundary, or whether they are not sure
whether a boundary is to be placed.\footnote{This could
be modelled as another boundary type, which would be
modelled in \metricacronym{} by the set of boundary types $T$.}
For this reason, we only characterize chance agreement between coders in
terms of one boundary presence category per type.

\subsubsection{Scott's~$\pi$}
Proposed by Scott~\shortcite{Scott:1955}, $\pi$ assumes that chance agreement
between coders can be characterized as the proportion of items that have been
assigned to category $k$ by both coders (Equation~\ref{eqn:scott_a_e}).
We calculate agreement ($\text{A}^\pi_a$) as pairwise mean
\metricacronym{} (scaled by each item's size) to enable agreement to
quantify near misses leniently, and chance agreement ($\text{A}^\pi_e$) can be calculated
as in Artstein and Poesio~\shortcite{ArtsteinPoesio:2008}.

\begin{footnotesize}
\spacersmall
\spacer
\begin{align}\label{eqn:scott_a_a}
\text{A}^\pi_a &= \frac{\sum_{i \in I} \text{mass}(i) \cdot
\text{S}(s_{i1},s_{i2})}
{\sum_{i \in I} \text{mass}(i)} \\
\label{eqn:scott_a_e} \text{A}^\pi_e &= \sum_{k \in K}
\big(\text{P}^\pi_e(k)\big)^2
\end{align}
\spacer
\spacersmall
\end{footnotesize}

We calculate chance agreement per category as the proportion of boundaries
($\text{seg}_t$) assigned by all coders over the total number of potential
boundaries for segmentations, as shown in Equation~\ref{eqn:scott_p_seg}.

\begin{footnotesize}
\spacer
\spacersmall
\begin{align}\label{eqn:scott_p_seg}
\text{P}^\pi_e(\text{seg}_t)   &= \frac{\sum_{c \in C}\sum_{i \in
I}|\text{boundaries}(t, s_{ic})|}
{\textbf{c} \cdot \sum_{i \in I} \big( \text{mass}(i) - 1 \big)}
\end{align}
\spacersmall
\spacer
\end{footnotesize}

This adapted coefficient appropriately estimates chance agreement in situations
where there no individual coder bias.

\subsubsection{Cohen's~$\kappa$}
Proposed by Cohen~\shortcite{Cohen:1960}, $\kappa$ characterizes chance
agreement as individual distributions per coder, calculated as shown in
Equations~\ref{eqn:cohen_a_a}-\ref{eqn:cohen_a_e} using our definition of
agreement ($\text{A}^\pi_a$) as shown earlier.

\begin{footnotesize}
\spacer
\spacer
\spacersmall
\spacersmall
\begin{align}\label{eqn:cohen_a_a}
\text{A}^\kappa_a &= \text{A}^\pi_a \\ \label{eqn:cohen_a_e}
\text{A}^\kappa_e &= \sum_{k \in K} \text{P}^\kappa_e(k|c_1) \cdot
\text{P}^\kappa_e(k|c_2)
\end{align}
\spacersmall
\spacersmall
\spacer
\end{footnotesize}

We calculate category probabilities as in Scott's $\pi$, but per coder, as shown
in Equation~\ref{eqn:cohen_pe_seg}.

\begin{footnotesize}
\spacer
\spacersmall
\begin{align}\label{eqn:cohen_pe_seg}
\text{P}^\kappa_e(\text{seg}_t|c)   &= \frac{\sum_{i \in
I}|\text{boundaries}(t, s_{ic})|}
{\sum_{i \in I} \big( \text{mass}(i) - 1 \big)}
\end{align}
\spacersmall
\spacer
\end{footnotesize}

This adapted coefficient appropriately estimates chance agreement for
segmentation evaluations where coder bias is present.

\subsubsection{Fleiss's~Multi-$\pi$}
Proposed by Fleiss~\shortcite{Fleiss:1971}, multi-$\pi$ ($\pi^*$) adapts
Scott's~$\pi$ for multiple annotators. We use
Artstein and~Poesio's~\shortcite[p. 564]{ArtsteinPoesio:2008} proposal for
calculating actual and expected agreement, and because all coders rate all
items, we express agreement as pairwise mean \metricacronym{} between all
coders as shown in Equations~\ref{eqn:mpo_a_a}-\ref{eqn:mpo_a_e}, adapting only
Equation~\ref{eqn:mpo_a_a}.

\begin{footnotesize}
\spacer
\begin{align}\label{eqn:mpo_a_a}
\text{A}^{\pi^*}_a &= \frac{1}{{\textbf{c} \choose 2}} \sum^{\textbf{c}-1}_{m=1}
\sum^{\textbf{c}}_{n=m+1} \frac{ \sum_{i \in I}
\text{mass}(i) \cdot \text{S}(s_{im},s_{in})}
{ \sum_{i \in I} \big( \text{mass}(i) - 1 \big)} \\
\label{eqn:mpo_a_e}
\text{A}^{\pi^*}_e &= \sum_{k \in K} \big(\text{P}^\pi_e(k)\big)^2
\end{align}
\spacer
\end{footnotesize}

\subsubsection{Fleiss's~Multi-$\kappa$}
Proposed by Davies~and~Fleiss~\shortcite{DaviesFleiss:1982}, multi-$\kappa$
($\kappa^*$) adapts Cohen's~$\kappa$ for multiple annotators.  We use
Artstein and Poesio's~\shortcite[extended version]{ArtsteinPoesio:2008} proposal
for calculating agreement just as in $\pi^*$, but with separate distributions
per coder as shown in Equations~\ref{eqn:mk_a_a}-\ref{eqn:mk_a_e}.

\begin{footnotesize}
\spacer
\begin{align}\label{eqn:mk_a_a}
\text{A}^{\kappa^*}_a &= \text{A}^{\pi^*}_a \\ \label{eqn:mk_a_e}
\text{A}^{\kappa^*}_e &= \sum_{k \in K} \bigg(
\frac{1}{{\textbf{c} \choose 2}} \sum^{\textbf{c}-1}_{m=1}
\sum^{\textbf{c}}_{n=m+1} \text{P}^\kappa_e(k|c_m) \cdot
\text{P}^\kappa_e(k|c_n) \bigg)
\end{align}
\spacer
\end{footnotesize}

\subsection{Annotator Bias}
To identify the degree of bias in a group of coders' segmentations, we can use a
measure of variance proposed by Artstein and Poesio~\shortcite[p.
572]{ArtsteinPoesio:2008} that is quantified in terms of the difference between
expected agreement when chance is assumed to vary between coders, and when it is
assumed to not.

\begin{footnotesize}
\spacer
\spacersmall
\spacersmall
\begin{align}\label{eqn:bias}
B = A^{\pi^*}_e - A^{\kappa^*}_e
\end{align}
\vspace{-1.9em}
\end{footnotesize}

\section{Experiments}\label{section:experiments}
To demonstrate the advantages of using \metricacronym{}, as opposed to
WindowDiff ($WD$), we compare both metrics using a variety of contrived
scenarios, and then compare our adapted agreement
coefficients against pairwise mean $WD$\footnote{Permuted, and with 
window size recalculated for each pair.} for the segmentations collected by
\KazantsevaSzpakowicz~\shortcite{KazantsevaSzpakowicz:2012}.

In this section, because $WD$ is a penalty-based metric, it is
reported as \oneminuswd{} so that it is easier to compare against \metricacronym{}
values. When reported in this way,  \oneminuswd{} and \metricacronym{} both range from
$[0,1]$, where 1 represents no errors and 0 represents maximal error.

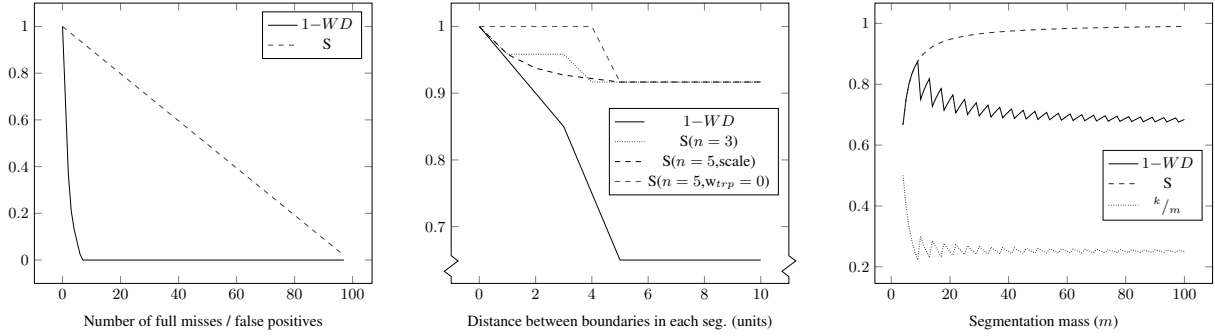
\begin{figure*}[t]
\spacer
\footnotesize
  \subfloat[Increasing the number of full misses, or FPs, where $k=25$ for $WD$]{
  \begin{tikzpicture}[scale=0.655000]
\begin{axis}[
xlabel = Number of full misses / false positives,
legend style={legend pos=north east}]
\addplot [color = black!100, mark = none] coordinates {
(0.0000,1.0000)
(1.0000,0.6842)
(2.0000,0.3684)
(3.0000,0.2105)
(4.0000,0.1316)
(5.0000,0.0789)
(6.0000,0.0263)
(7.0000,0.0000)
(8.0000,0.0000)
(9.0000,0.0000)
(10.0000,0.0000)
(11.0000,0.0000)
(12.0000,0.0000)
(13.0000,0.0000)
(14.0000,0.0000)
(15.0000,0.0000)
(16.0000,0.0000)
(17.0000,0.0000)
(18.0000,0.0000)
(19.0000,0.0000)
(20.0000,0.0000)
(21.0000,0.0000)
(22.0000,0.0000)
(23.0000,0.0000)
(24.0000,0.0000)
(25.0000,0.0000)
(26.0000,0.0000)
(27.0000,0.0000)
(28.0000,0.0000)
(29.0000,0.0000)
(30.0000,0.0000)
(31.0000,0.0000)
(32.0000,0.0000)
(33.0000,0.0000)
(34.0000,0.0000)
(35.0000,0.0000)
(36.0000,0.0000)
(37.0000,0.0000)
(38.0000,0.0000)
(39.0000,0.0000)
(40.0000,0.0000)
(41.0000,0.0000)
(42.0000,0.0000)
(43.0000,0.0000)
(44.0000,0.0000)
(45.0000,0.0000)
(46.0000,0.0000)
(47.0000,0.0000)
(48.0000,0.0000)
(49.0000,0.0000)
(50.0000,0.0000)
(51.0000,0.0000)
(52.0000,0.0000)
(53.0000,0.0000)
(54.0000,0.0000)
(55.0000,0.0000)
(56.0000,0.0000)
(57.0000,0.0000)
(58.0000,0.0000)
(59.0000,0.0000)
(60.0000,0.0000)
(61.0000,0.0000)
(62.0000,0.0000)
(63.0000,0.0000)
(64.0000,0.0000)
(65.0000,0.0000)
(66.0000,0.0000)
(67.0000,0.0000)
(68.0000,0.0000)
(69.0000,0.0000)
(70.0000,0.0000)
(71.0000,0.0000)
(72.0000,0.0000)
(73.0000,0.0000)
(74.0000,0.0000)
(75.0000,0.0000)
(76.0000,0.0000)
(77.0000,0.0000)
(78.0000,0.0000)
(79.0000,0.0000)
(80.0000,0.0000)
(81.0000,0.0000)
(82.0000,0.0000)
(83.0000,0.0000)
(84.0000,0.0000)
(85.0000,0.0000)
(86.0000,0.0000)
(87.0000,0.0000)
(88.0000,0.0000)
(89.0000,0.0000)
(90.0000,0.0000)
(91.0000,0.0000)
(92.0000,0.0000)
(93.0000,0.0000)
(94.0000,0.0000)
(95.0000,0.0000)
(96.0000,0.0000)
(97.0000,0.0000)
};
\addlegendentry{\oneminuswd{}};
\addplot [color = black!75, dashed, mark = none] coordinates {
(0.0000,1.0000)
(1.0000,0.9899)
(2.0000,0.9798)
(3.0000,0.9697)
(4.0000,0.9596)
(5.0000,0.9495)
(6.0000,0.9394)
(7.0000,0.9293)
(8.0000,0.9192)
(9.0000,0.9091)
(10.0000,0.8990)
(11.0000,0.8889)
(12.0000,0.8788)
(13.0000,0.8687)
(14.0000,0.8586)
(15.0000,0.8485)
(16.0000,0.8384)
(17.0000,0.8283)
(18.0000,0.8182)
(19.0000,0.8081)
(20.0000,0.7980)
(21.0000,0.7879)
(22.0000,0.7778)
(23.0000,0.7677)
(24.0000,0.7576)
(25.0000,0.7475)
(26.0000,0.7374)
(27.0000,0.7273)
(28.0000,0.7172)
(29.0000,0.7071)
(30.0000,0.6970)
(31.0000,0.6869)
(32.0000,0.6768)
(33.0000,0.6667)
(34.0000,0.6566)
(35.0000,0.6465)
(36.0000,0.6364)
(37.0000,0.6263)
(38.0000,0.6162)
(39.0000,0.6061)
(40.0000,0.5960)
(41.0000,0.5859)
(42.0000,0.5758)
(43.0000,0.5657)
(44.0000,0.5556)
(45.0000,0.5455)
(46.0000,0.5354)
(47.0000,0.5253)
(48.0000,0.5152)
(49.0000,0.5051)
(50.0000,0.4949)
(51.0000,0.4848)
(52.0000,0.4747)
(53.0000,0.4646)
(54.0000,0.4545)
(55.0000,0.4444)
(56.0000,0.4343)
(57.0000,0.4242)
(58.0000,0.4141)
(59.0000,0.4040)
(60.0000,0.3939)
(61.0000,0.3838)
(62.0000,0.3737)
(63.0000,0.3636)
(64.0000,0.3535)
(65.0000,0.3434)
(66.0000,0.3333)
(67.0000,0.3232)
(68.0000,0.3131)
(69.0000,0.3030)
(70.0000,0.2929)
(71.0000,0.2828)
(72.0000,0.2727)
(73.0000,0.2626)
(74.0000,0.2525)
(75.0000,0.2424)
(76.0000,0.2323)
(77.0000,0.2222)
(78.0000,0.2121)
(79.0000,0.2020)
(80.0000,0.1919)
(81.0000,0.1818)
(82.0000,0.1717)
(83.0000,0.1616)
(84.0000,0.1515)
(85.0000,0.1414)
(86.0000,0.1313)
(87.0000,0.1212)
(88.0000,0.1111)
(89.0000,0.1010)
(90.0000,0.0909)
(91.0000,0.0808)
(92.0000,0.0707)
(93.0000,0.0606)
(94.0000,0.0505)
(95.0000,0.0404)
(96.0000,0.0303)
(97.0000,0.0202)
};
\addlegendentry{S};
\end{axis}
\end{tikzpicture}
  \label{fig:increasing.misses}
  }
  \hspace{1em}
  \subfloat[Increasing the distance between two boundaries considered to be a near miss until metrics consider them a full miss]{
  \begin{tikzpicture}[scale=0.655]
\begin{axis}[
xlabel = Distance between boundaries in each seg. (units),
axis y discontinuity=crunch,
legend style={at={(0.97,0.47)}, anchor=east}]
\addplot [color = black!100, mark = none] coordinates {
(0.0000,1.0000)
(1.0000,0.9500)
(2.0000,0.9000)
(3.0000,0.8500)
(4.0000,0.7500)
(5.0000,0.6500)
(6.0000,0.6500)
(7.0000,0.6500)
(8.0000,0.6500)
(9.0000,0.6500)
(10.0000,0.6500)
};
\addlegendentry{\oneminuswd{}};
\addplot [color = black!100, densely dotted, mark = none] coordinates {
(0.0000,1.0000)
(1.0000,0.9583)
(2.0000,0.9583)
(3.0000,0.9583)
(4.0000,0.9167)
(5.0000,0.9167)
(6.0000,0.9167)
(7.0000,0.9167)
(8.0000,0.9167)
(9.0000,0.9167)
(10.0000,0.9167)
};
\addlegendentry{S($n=3$)};
\addplot [color = black, dashed, mark = none] coordinates {
(0.0000,1.0000)
(1.0000,0.9583)
(2.0000,0.9375)
(3.0000,0.9271)
(4.0000,0.9219)
(5.0000,0.9167)
(6.0000,0.9167)
(7.0000,0.9167)
(8.0000,0.9167)
(9.0000,0.9167)
(10.0000,0.9167)
};
\addlegendentry{S($n=5$,scale)};
\addplot [color = black!70, dashed, mark = none] coordinates {
(0.0000,1.0000)
(1.0000,1.0000)
(2.0000,1.0000)
(3.0000,1.0000)
(4.0000,1.0000)
(5.0000,0.9167)
(6.0000,0.9167)
(7.0000,0.9167)
(8.0000,0.9167)
(9.0000,0.9167)
(10.0000,0.9167)
};
\addlegendentry{S($n=5$,$\text{w}_{trp}=0$)};
\end{axis}
\end{tikzpicture}
  \label{fig:increasing.transp.dist}
  }
  \hspace{1em}
  \subfloat[Increasing the mass $m$ of segmentations configured as shown in Figure~\ref{fig:increasing.size.segs} showing the effect of $k$ on \oneminuswd{}]{
  \begin{tikzpicture}[scale=0.655000]
\begin{axis}[
xlabel = Segmentation mass ($m$),
legend style={at={(0.97,0.35)}, anchor=east}]
\addplot [color = black!100, mark = none] coordinates {
(4.0000,0.6667)
(5.0000,0.7500)
(6.0000,0.8000)
(7.0000,0.8333)
(8.0000,0.8571)
(9.0000,0.8750)
(10.0000,0.7500)
(11.0000,0.7778)
(12.0000,0.8000)
(13.0000,0.8182)
(14.0000,0.7273)
(15.0000,0.7500)
(16.0000,0.7692)
(17.0000,0.7857)
(18.0000,0.7143)
(19.0000,0.7333)
(20.0000,0.7500)
(21.0000,0.7647)
(22.0000,0.7059)
(23.0000,0.7222)
(24.0000,0.7368)
(25.0000,0.7500)
(26.0000,0.7000)
(27.0000,0.7143)
(28.0000,0.7273)
(29.0000,0.7391)
(30.0000,0.6957)
(31.0000,0.7083)
(32.0000,0.7200)
(33.0000,0.7308)
(34.0000,0.6923)
(35.0000,0.7037)
(36.0000,0.7143)
(37.0000,0.7241)
(38.0000,0.6897)
(39.0000,0.7000)
(40.0000,0.7097)
(41.0000,0.7188)
(42.0000,0.6875)
(43.0000,0.6970)
(44.0000,0.7059)
(45.0000,0.7143)
(46.0000,0.6857)
(47.0000,0.6944)
(48.0000,0.7027)
(49.0000,0.7105)
(50.0000,0.6842)
(51.0000,0.6923)
(52.0000,0.7000)
(53.0000,0.7073)
(54.0000,0.6829)
(55.0000,0.6905)
(56.0000,0.6977)
(57.0000,0.7045)
(58.0000,0.6818)
(59.0000,0.6889)
(60.0000,0.6957)
(61.0000,0.7021)
(62.0000,0.6809)
(63.0000,0.6875)
(64.0000,0.6939)
(65.0000,0.7000)
(66.0000,0.6800)
(67.0000,0.6863)
(68.0000,0.6923)
(69.0000,0.6981)
(70.0000,0.6792)
(71.0000,0.6852)
(72.0000,0.6909)
(73.0000,0.6964)
(74.0000,0.6786)
(75.0000,0.6842)
(76.0000,0.6897)
(77.0000,0.6949)
(78.0000,0.6780)
(79.0000,0.6833)
(80.0000,0.6885)
(81.0000,0.6935)
(82.0000,0.6774)
(83.0000,0.6825)
(84.0000,0.6875)
(85.0000,0.6923)
(86.0000,0.6769)
(87.0000,0.6818)
(88.0000,0.6866)
(89.0000,0.6912)
(90.0000,0.6765)
(91.0000,0.6812)
(92.0000,0.6857)
(93.0000,0.6901)
(94.0000,0.6761)
(95.0000,0.6806)
(96.0000,0.6849)
(97.0000,0.6892)
(98.0000,0.6757)
(99.0000,0.6800)
(100.0000,0.6842)
};
\addlegendentry{\oneminuswd{}};
\addplot [color = black!83, dashed, mark = none] coordinates {
(4.0000,0.6667)
(5.0000,0.7500)
(6.0000,0.8000)
(7.0000,0.8333)
(8.0000,0.8571)
(9.0000,0.8750)
(10.0000,0.8889)
(11.0000,0.9000)
(12.0000,0.9091)
(13.0000,0.9167)
(14.0000,0.9231)
(15.0000,0.9286)
(16.0000,0.9333)
(17.0000,0.9375)
(18.0000,0.9412)
(19.0000,0.9444)
(20.0000,0.9474)
(21.0000,0.9500)
(22.0000,0.9524)
(23.0000,0.9545)
(24.0000,0.9565)
(25.0000,0.9583)
(26.0000,0.9600)
(27.0000,0.9615)
(28.0000,0.9630)
(29.0000,0.9643)
(30.0000,0.9655)
(31.0000,0.9667)
(32.0000,0.9677)
(33.0000,0.9688)
(34.0000,0.9697)
(35.0000,0.9706)
(36.0000,0.9714)
(37.0000,0.9722)
(38.0000,0.9730)
(39.0000,0.9737)
(40.0000,0.9744)
(41.0000,0.9750)
(42.0000,0.9756)
(43.0000,0.9762)
(44.0000,0.9767)
(45.0000,0.9773)
(46.0000,0.9778)
(47.0000,0.9783)
(48.0000,0.9787)
(49.0000,0.9792)
(50.0000,0.9796)
(51.0000,0.9800)
(52.0000,0.9804)
(53.0000,0.9808)
(54.0000,0.9811)
(55.0000,0.9815)
(56.0000,0.9818)
(57.0000,0.9821)
(58.0000,0.9825)
(59.0000,0.9828)
(60.0000,0.9831)
(61.0000,0.9833)
(62.0000,0.9836)
(63.0000,0.9839)
(64.0000,0.9841)
(65.0000,0.9844)
(66.0000,0.9846)
(67.0000,0.9848)
(68.0000,0.9851)
(69.0000,0.9853)
(70.0000,0.9855)
(71.0000,0.9857)
(72.0000,0.9859)
(73.0000,0.9861)
(74.0000,0.9863)
(75.0000,0.9865)
(76.0000,0.9867)
(77.0000,0.9868)
(78.0000,0.9870)
(79.0000,0.9872)
(80.0000,0.9873)
(81.0000,0.9875)
(82.0000,0.9877)
(83.0000,0.9878)
(84.0000,0.9880)
(85.0000,0.9881)
(86.0000,0.9882)
(87.0000,0.9884)
(88.0000,0.9885)
(89.0000,0.9886)
(90.0000,0.9888)
(91.0000,0.9889)
(92.0000,0.9890)
(93.0000,0.9891)
(94.0000,0.9892)
(95.0000,0.9894)
(96.0000,0.9895)
(97.0000,0.9896)
(98.0000,0.9897)
(99.0000,0.9898)
(100.0000,0.9899)
};
\addlegendentry{S};
\addplot [color = black!100, densely dotted, mark = none] coordinates {
(4.0000,0.5000)
(5.0000,0.4000)
(6.0000,0.3333)
(7.0000,0.2857)
(8.0000,0.2500)
(9.0000,0.2222)
(10.0000,0.3000)
(11.0000,0.2727)
(12.0000,0.2500)
(13.0000,0.2308)
(14.0000,0.2857)
(15.0000,0.2667)
(16.0000,0.2500)
(17.0000,0.2353)
(18.0000,0.2778)
(19.0000,0.2632)
(20.0000,0.2500)
(21.0000,0.2381)
(22.0000,0.2727)
(23.0000,0.2609)
(24.0000,0.2500)
(25.0000,0.2400)
(26.0000,0.2692)
(27.0000,0.2593)
(28.0000,0.2500)
(29.0000,0.2414)
(30.0000,0.2667)
(31.0000,0.2581)
(32.0000,0.2500)
(33.0000,0.2424)
(34.0000,0.2647)
(35.0000,0.2571)
(36.0000,0.2500)
(37.0000,0.2432)
(38.0000,0.2632)
(39.0000,0.2564)
(40.0000,0.2500)
(41.0000,0.2439)
(42.0000,0.2619)
(43.0000,0.2558)
(44.0000,0.2500)
(45.0000,0.2444)
(46.0000,0.2609)
(47.0000,0.2553)
(48.0000,0.2500)
(49.0000,0.2449)
(50.0000,0.2600)
(51.0000,0.2549)
(52.0000,0.2500)
(53.0000,0.2453)
(54.0000,0.2593)
(55.0000,0.2545)
(56.0000,0.2500)
(57.0000,0.2456)
(58.0000,0.2586)
(59.0000,0.2542)
(60.0000,0.2500)
(61.0000,0.2459)
(62.0000,0.2581)
(63.0000,0.2540)
(64.0000,0.2500)
(65.0000,0.2462)
(66.0000,0.2576)
(67.0000,0.2537)
(68.0000,0.2500)
(69.0000,0.2464)
(70.0000,0.2571)
(71.0000,0.2535)
(72.0000,0.2500)
(73.0000,0.2466)
(74.0000,0.2568)
(75.0000,0.2533)
(76.0000,0.2500)
(77.0000,0.2468)
(78.0000,0.2564)
(79.0000,0.2532)
(80.0000,0.2500)
(81.0000,0.2469)
(82.0000,0.2561)
(83.0000,0.2530)
(84.0000,0.2500)
(85.0000,0.2471)
(86.0000,0.2558)
(87.0000,0.2529)
(88.0000,0.2500)
(89.0000,0.2472)
(90.0000,0.2556)
(91.0000,0.2527)
(92.0000,0.2500)
(93.0000,0.2473)
(94.0000,0.2553)
(95.0000,0.2526)
(96.0000,0.2500)
(97.0000,0.2474)
(98.0000,0.2551)
(99.0000,0.2525)
(100.0000,0.2500)
};
\addlegendentry{$^k/_m$};
\end{axis}
\end{tikzpicture}
  \label{fig:increasing.size}
  }
  \caption{Responses of \oneminuswd{} and S to various segmentation scenarios}
\spacer
\spacer
\end{figure*}

\subsection{Segmentation Cases}
%AK:  you know, my general feeling about this section is that it was written in
% haste. Your experiments are all proper and convincing, your graphs are very
% nice but the narrative is somehow not smooth nor convincing.
\subsubsection*{}
\vspace{-1.5em}

\paragraph{Maximal versus minimal segmentation}  When proposing a new metric,
its reactions to extrema must be illustrated, for example when a maximal
segmentation is compared to a minimal segmentation, as shown in
Figure~\ref{fig:maximal.minimal}.  In this scenario, both \oneminuswd{} and
\metricacronym{} appropriately identify that this case represents
maximal error, or 0. Though not shown here, both metrics also report a
similarity of 1.0 when identical segmentations are compared.

\begin{figure}[h]
% [14]
% [1,1,1,1,1,1,1,1,1,1,1,1,1,1]
  \centering
\begin{tikzpicture}[node distance=0cm, outer sep=0pt, font=\footnotesize,
scale=0.7]
\node[header] (ref) at (1,8) 		 {\scriptsize{$s_1$}};
\node[header] (hyp) [below = of ref] {\scriptsize{$s_2$}};
\node[block14](r1)  [right = of ref] {14};
\node[block1] (h1)  [right = of hyp] {1};
\node[block1] (h2)  [right = of h1]  {1};
\node[block1] (h3)  [right = of h2]  {1};
\node[block1] (h4)  [right = of h3]  {1};
\node[block1] (h5)  [right = of h4]  {1};
\node[block1] (h6)  [right = of h5]  {1};
\node[block1] (h7)  [right = of h6]  {1};
\node[block1] (h8)  [right = of h7]  {1};
\node[block1] (h9)  [right = of h8]  {1};
\node[block1] (h10) [right = of h9]  {1};
\node[block1] (h11) [right = of h10]  {1};
\node[block1] (h12) [right = of h11]  {1};
\node[block1] (h13) [right = of h12]  {1};
\node[block1] (h14) [right = of h13]  {1};
\end{tikzpicture}
  \caption{Maximal versus minimal seg. masses}
  \label{fig:maximal.minimal}
  \spacer
  \spacer
\end{figure}

\paragraph{Full misses}  For the most serious source of error, full misses (\ie
FPs and FNs), both metrics appropriately report a reduction in similarity for
cases such as Figure~\ref{fig:full.misses} that is very similar (\oneminuswd{} $ =
0.8462$, \metricacronym{}$ = 0.8461$). Where the two metrics differ is when this
type of error is increased.

\begin{figure}[h]
% [1,2,2,2,4,2,1]
% [1,2,8,    2,1]
  \centering
\begin{tikzpicture}[node distance=0cm, outer sep=0pt, font=\footnotesize,
scale=0.7]
\node[header] (ref) at (1,8) 		 {\scriptsize{$s_1$}};
\node[header] (hyp) [below = of ref] {\scriptsize{$s_2$}};
\node[block1] (r1)  [right = of ref] {1};
\node[block2] (r2)  [right = of r1]  {2};
\node[block2] (r3)  [right = of r2]  {2};
\node[block2] (r4)  [right = of r3]  {2};
\node[block4] (r5)  [right = of r4]  {4};
\node[block2] (r6)  [right = of r5]  {2};
\node[block1] (r7)  [right = of r6]  {1};
\node[block1] (h1)  [right = of hyp] {1};
\node[block2] (h2)  [right = of h1]  {2};
\node[block8] (h3)  [right = of h2]  {8};
\node[block2] (h4)  [right = of h3]  {2};
\node[block1] (h5)  [right = of h4]  {1};
\end{tikzpicture}
  \caption{Full misses in seg. masses}
  \label{fig:full.misses}
  \spacer
\end{figure}

\metricacronym{} reacts to increasing full misses linearly, whereas WindowDiff
can prematurely report a maximal number of errors.
Figure~\ref{fig:increasing.misses} demonstrates this effect, where for each
iteration we have taken segmentations of 100 \sizeunits{} of mass with one
matching boundary at the first hypothesis boundary position, and uniformly
increased the number of internal hypothesis segments, giving us 1 matching
boundary, and $[0,98]$ FPs.  This premature report of maximal error (at 7 FP) by $WD$ is
caused by the window size ($k=25$) being greater than all of the internal
hypothesis segment sizes, making all windows penalized for containing errors.

\begin{table*}[t]
\scriptsize
\centering
\setlength{\tabcolsep}{0.6em}
\begin{tabular}{ c c c p{0.1em} c c p{0.1em} c c }
 & \multicolumn{2}{c}{\textbf{Scenario 1: FN, $p=0.5$}} 
 & & \multicolumn{2}{c}{\textbf{Scenario 2: FP, $p=0.5$}} 
 & & \multicolumn{2}{c}{\textbf{Scenario 3: FP and FN, $p=0.5$}} \\ \hline
 & (20,30) & (15,35) & & (20,30) & (15,35) & & (20,30) & (15,35) \\ \hline
 $WD$ &
$0.2340\pm0.0113$ &
$0.2292\pm0.0104$ & &
$0.2265\pm0.0114$ &
$0.2265\pm0.0111$ & &
$0.3635\pm0.0126$ &
$0.3599\pm0.0117$ \\
 S &
$0.9801\pm0.0006$ &
$0.9801\pm0.0006$ & &
$0.9800\pm0.0006$ &
$0.9800\pm0.0006$ & &
$0.9605\pm0.0009$ &
$0.9603\pm0.0009$ \\ \hline
 & (10,40) & (5,45) & & (10,40) & (5,45) & & (10,40) & (5,45)  \\ \hline
 $WD$ &
$0.2297\pm0.0105$ &
$0.2206\pm0.0079$ & &
$0.2256\pm0.0102$ &
$0.2184\pm0.0069$ & &
$0.3516\pm0.0110$ &
$0.3254\pm0.0087$ \\
 S &
$0.9799\pm0.0007$ &
$0.9796\pm0.0007$ & &
$0.9800\pm0.0006$ &
$0.9796\pm0.0007$ & &
$0.9606\pm0.0010$ &
$0.9598\pm0.0011$ 
\end{tabular}

\spacer
\caption{Stability of mean (with standard deviation) values of $WD$ and S
in three different scenarios, each defining the: probability of a false positive
(FP), false negative (FN), or both.  Each scenario varies the range of
internal segment sizes (\eg $(20,30)$). Low standard deviation and similar
within-scenario means demonstrates low sensitivity to variations in internal segment size.}
\label{table:variation}
\spacer
\spacer
\spacer
\end{table*}

\paragraph{Near misses}  When dealing with near misses, the values of both
metrics drop (\oneminuswd{} $ = 0.8182$, $\text{\metricacronym{}} = 0.9231$), but to
greatly varying degrees.  In comparison to full misses, WindowDiff penalizes a
near miss, like that in Figure~\ref{fig:near.miss.1}, far more than
\metricacronym{}. This difference is due to the distance between the two
boundaries involved in a near miss; \metricacronym{} shows, in this case, 1
\errorunit{} of error until it is outside of the $n$-wise transposition window
(where $n=2$ \errorunits{}), at which point it is considered an error of not one
transposition, but two substitutions (2 \errorunits{}).

\begin{figure}[h]
% [6,8]
% [7,7]
  \centering
\begin{tikzpicture}[node distance=0cm, outer sep=0pt, font=\footnotesize,
scale=0.7]
\node[header] (ref) at (1,8) 		 {\scriptsize{$s_1$}};
\node[header] (hyp) [below = of ref] {\scriptsize{$s_2$}};
\node[block6] (r1)  [right = of ref] {6};
\node[block8] (r2)  [right = of r1]  {8};
\node[block7] (h1)  [right = of hyp] {7};
\node[block7] (h2)  [right = of h1]  {7};
\end{tikzpicture}
\spacer
  \caption{Near misses in seg. masses}
  \label{fig:near.miss.1}
\spacer
\spacersmall
\end{figure}

If we wanted to completely forgive near misses up to $n$ \errorunits{}, we could
set the weighting of transpositions in \metricacronym{} to $w_{trp}=0$.  This is
useful if a specific segmentation task accepts that near misses are very
probable, and that there is little cost associated with a near miss in a window
of $n$ \errorunits{}. We can also set $n$ to a high number, \ie 5 \errorunits{}, and
use the scaled transposition error (te) function
(Equation~\ref{eqn:transposition.penalty}) to slowly increase the error from
$b=1$ \errorunit{} to $b=2$ \errorunits{}, as shown in
Figure~\ref{fig:increasing.transp.dist}, which shows how both metrics react to
increases in the distance between a near miss in a segment of 25 \sizeunits{}.
These configurations are all preferable to the drop of \oneminuswd{}.

\subsection{Segmentation Mass Scale Effects}
It is important for a segmentation evaluation metric to take into account the
severity of an error in terms of segment size.  An error in a 100 \sizeunit{}
segment should be considered less severe than an error in a 2 \sizeunit{}
segment, because an extra boundary placed within a 100 \sizeunit{} segment 
(\eg Figure~\ref{fig:small.and.large.segs} with $m=100$)
could probably indicate a weak boundary, whereas in a 4 \sizeunit{} segment the
probability that an extra boundary exists right next to two agreed-upon boundaries
should be small for most tasks, meaning that it is probable that the extra
boundary is an error, and not a weak boundary.

\begin{figure}[h]
% [6,8]
% [7,7]
  \centering
\spacersmall
\resizebox{0.49\textwidth}{!}{
\begin{tikzpicture}[node distance=0cm, outer sep=0pt, font=\footnotesize,
scale=0.7]
\node[header] (ref) at (1,8) 		 {\scriptsize{$s_1$}};
\node[header] (hyp) [below = of ref] {\scriptsize{$s_2$}};
\node[block4]  (r1)  [right = of ref] {\tiny{$^{m}/_{4}$}};
\node[block8] (r2)  [right = of r1]  {\tiny{$^{m}/_{2}$}};
\node[block4] (r3)  [right = of r2]  {\tiny{$^{m}/_{4}$}};
\node[block4]  (h1)  [right = of hyp] {\tiny{$^{m}/_{4}$}};
\node[block4]  (h2)  [right = of h1]  {\tiny{$^{m}/_{4}$}};
\node[block4]  (h3)  [right = of h2]  {\tiny{$^{m}/_{4}$}};
\node[block4]  (h4)  [right = of h3]  {\tiny{$^{m}/_{4}$}};
\end{tikzpicture}}
\spacer
\spacer
\spacer
\spacer
  \caption{Two segmentations of mass $m$ with a full miss}
  \label{fig:small.and.large.segs}
\spacer
\end{figure}

To demonstrate that \metricacronym{} is sensitive to segment size,
Figure~\ref{fig:increasing.size} shows how \metricacronym{} and \oneminuswd{} respond
when comparing segmentations configured as shown in
Figure~\ref{fig:increasing.size.segs} (containing one match and one full miss)
with linearly increasing mass ($4 \leq m \leq 100$).
\oneminuswd{} will eventually indicate 0.68, whereas \metricacronym{}
appropriately discounts the error as mass is increased, approaching 1 as
$\lim_{m \to \infty}$.  \oneminuswd{} behaves in this way because of how it calculates
its window size parameter, $k$, which is plotted as $k/m$ to show how its value
influences \oneminuswd{}.

\begin{figure}[h]
% [6,8]
% [7,7]
  \centering
\begin{tikzpicture}[node distance=0cm, outer sep=0pt, font=\footnotesize,
scale=0.7]
\node[header] (ref) at (1,8) 		 {\scriptsize{$s_1$}};
\node[header] (hyp) [below = of ref] {\scriptsize{$s_2$}};
\node[block4]  (r1)  [right = of ref] {\tiny{$^{m}/_{4}$}};
\node[block11] (r2)  [right = of r1]  {\tiny{$m - (^{m}/_{4})$}};
\node[block4]  (h1)  [right = of hyp] {\tiny{$^{m}/_{4}$}};
\node[block4]  (h2)  [right = of h1]  {\tiny{$^{m}/_{4}$}};
\node[block7]  (h3)  [right = of h2]  {\tiny{$^{m}/_{2}$}};
\end{tikzpicture}
\spacer
  \caption{Two segmentations of mass $m$ compared with increasing $m$ in
  Figure~\ref{fig:increasing.size} ($s_1$ as reference)}
  \label{fig:increasing.size.segs}
\spacer
\spacer
\spacer
\spacersmall
\end{figure}

\subsection{Variation in Segment Sizes}
When Pevzner~and~Hearst~\shortcite{PevznerHearst:2002} proposed $WD$, they
demonstrated that it was not as sensitive as \Pk{} to variations in the size of
segments inside a segmentation.  To show this, they simulated how $WD$
performs upon a segmentation comprised of 1000 segments with four different
uniformly distributed ranges of internal segment sizes (keeping the mean at
approximately 25 \sizeunits{}) in comparison to a hypothesis segmentation with
errors (false positives, false negatives, and both) uniformly distributed within
segments \cite[pp. 11--12]{PevznerHearst:2002}.  10 trials were performed for
each segment size range and error probability, with 100 hypotheses generated per
trial.  Recreating this simulation, we compare the stability of \metricacronym{}
in comparison to $WD$, as shown in Table~\ref{table:variation}. We can see
that $WD$ values show substantial within-scenario variation for each segment
size range, and larger standard deviations, than \metricacronym{}.

\subsection{Inter-Annotator Agreement Coefficients}
\begin{figure}[t]
\spacer
\spacersmall
  \centering
  \begin{tikzpicture}[inner sep=0pt,outer sep=0pt,node distance=0em,
  font=\footnotesize]
  % Heatmap
  \input{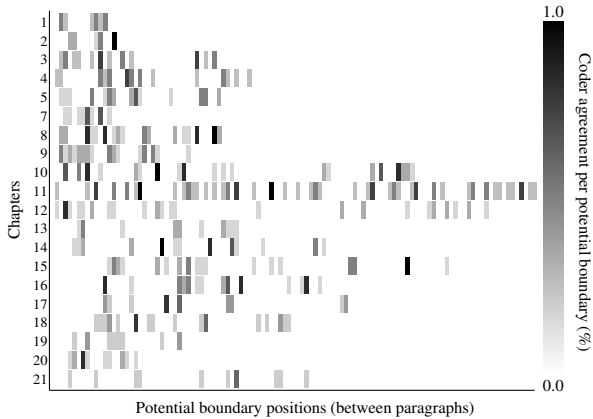}
  % Legend
  \node[draw=none, top color=black,bottom color=white, text height=17.5em,
  text width=1em, scale=0.7] (legend) at
  (6.8,-2.35) {};
  \node[draw=none, scale=0.7, above of = legend, node distance=9.25em] (l1)
  {1.0};
  \node[draw=none, scale=0.7, below of = legend, node distance=9.25em] (l0)
  {0.0};
  % Axis labels
  \node[rotate=90, scale=0.7] (chapters) at (-0.35,-2.5)
  {Chapters};
  \node[scale=0.7] (boundaries) at (3.5,-5.15)
  {Potential boundary positions (between paragraphs)};
  \node[rotate=-90, scale=0.7] (agreement) at (7.2,-2.4)
  {Coder agreement per potential boundary (\%)};
  % Axis
  \draw (0.1,-4.90) -- coordinate (x axis mid) (6.55,-4.90);
  \draw (0.1, 0.15) -- coordinate (y axis mid) (0.1, -4.90);
  % End
  \end{tikzpicture}
\spacer
  \caption{Heat maps for the segmentations of each chapter showing the
  percentage of coders who agree upon boundary positions (darker shows higher agreement)}
  \label{fig:segmentation.heatmaps}
\spacersmall
\spacer
\spacer
\end{figure}

Here, we demonstrate the adapted inter-annotator agreement coefficients
upon topical paragraph-level segmentations produced by 27 coders of 20 chapters from the novel
\emph{The Moonstone} by Wilkie Collins collected by
\KazantsevaSzpakowicz~\shortcite{KazantsevaSzpakowicz:2012}.
Figure~\ref{fig:segmentation.heatmaps} shows a heat map of each chapter where
the percentage of coders who agreed upon each potential boundary is represented.
Comparing this heat map to the inter-annotator agreement coefficients in
Table~\ref{table:agreements} allows us to better understand why certain chapters
have lower reliability.

% Low score
Chapter 1 has the lowest $\pi^*_{S}$ score in the table, and also the highest
bias ($B_{S}$).  One of the reasons for this low reliability can be attributed
to the chapter's small mass ($m$) and few coders ($|c|$), which makes it more
sensitive to chance agreement. Visually, the predominance of grey indicates
that, although there are probably two boundaries, their exact location is not
very well agreed upon.  In this case, \oneminuswd{} incorrectly indicates the opposite,
that this chapter may have relatively moderate reliability, because it is not
corrected for chance agreement.

\oneminuswd{} indicates that the lowest reliability is found in Chapter 19.
$\pi^*_{S}$ indicates that this is one of the higher agreement chapters, and
looking at the heat map, we can see that it does not contain any strongly agreed
upon boundaries.  In this chapter, there is little opportunity to agree by
chance due to the low number of boundaries ($|b|$) placed, and because the
judgements are tightly clustered in a fair amount of mass, the \metricacronym{}
component of $\pi^*_{S}$ appropriately takes into account the near misses
observed and gives it a high reliability score.

% High score
Chapter 17 received the highest $\pi^*_{S}$ in the table, which is another
example of how tight clustering of boundary choices in a large mass leads
$\pi^*_{S}$ to appropriately indicate high reliability despite that there are
not as many individual highly-agreed-upon boundaries, whereas \oneminuswd{} indicates
that there is low reliability.  \oneminuswd{} and $\pi^*_{S}$ both agree, however, that
chapter 16 has high reliability.

% Summary
Despite WindowDiff's sensitivity to near misses, it is evident that its pairwise
mean cannot be used to consistently judge inter-annotator agreement, or
reliability.  \metricacronym{} demonstrates better versatility when accounting
for near misses, and when used as part of inter-annotator agreement
coefficients, it properly takes into account chance agreement.  Following
Artstein~and~Poesio's~\shortcite[pp. 590--591]{ArtsteinPoesio:2008}
recommendation, and given the low bias (mean coder group $B_{S} =
0.0061\pm0.0035$), we propose reporting reliability using $\pi^*$
for this corpus, where the mean coder group $\pi^*_{S}$ for the corpus is
$0.8904\pm0.0392$ (counting 1039 full and 212 near misses).

\begin{table}[t]
\spacer
\spacersmall
  \scriptsize
  \setlength{\tabcolsep}{0.5em}
    \begin{tabular}{ r r r r r r r r r }
  \multicolumn{1}{c}{\emph{Ch.}} & 
  \multicolumn{1}{c}{\emph{$\pi^*_{S}$}} & 
  \multicolumn{1}{c}{\emph{$\kappa^*_{S}$}} & 
  \multicolumn{1}{c}{\emph{$B_{S}$}} & 
  \multicolumn{1}{c}{\emph{\oneminuswd{}}} & 
  \multicolumn{1}{c}{\emph{$|c|$}} & 
  \multicolumn{1}{c}{\emph{$|b|$}} & 
  \multicolumn{1}{c}{\emph{m}} \\ \hline
1	& 0.7452	& 0.7463	& 0.0039	& $0.6641\pm0.1307$	& 4	& 13	& 13\\
2	& 0.8839	& 0.8840	& 0.0009	& $0.7619\pm0.1743$	& 6	& 20	& 15\\
3	& 0.8338	& 0.8340	& 0.0013	& $0.6732\pm0.1559$	& 4	& 23	& 38\\
4	& 0.8414	& 0.8417	& 0.0019	& $0.6019\pm0.2245$	& 4	& 25	& 46\\
5	& 0.8773	& 0.8774	& 0.0003	& $0.6965\pm0.1106$	& 6	& 34	& 42\\
7	& 0.8132	& 0.8133	& 0.0002	& $0.6945\pm0.1822$	& 6	& 20	& 15\\
8	& 0.8495	& 0.8496	& 0.0006	& $0.7505\pm0.0911$	& 6	& 48	& 39\\
9	& 0.8104	& 0.8105	& 0.0009	& $0.6502\pm0.1319$	& 6	& 35	& 33\\
10	& 0.9077	& 0.9078	& 0.0002	& $0.7729\pm0.0770$	& 6	& 56	& 83\\
11	& 0.8130	& 0.8135	& 0.0022	& $0.6189\pm0.1294$	& 4	& 73	& 111\\
12	& 0.9178	& 0.9178	& 0.0001	& $0.6504\pm0.1277$	& 6	& 40	& 102\\
13	& 0.9354	& 0.9354	& 0.0002	& $0.5660\pm0.2187$	& 6	& 21	& 58\\
14	& 0.9367	& 0.9367	& 0.0001	& $0.7128\pm0.1744$	& 6	& 35	& 70\\
15	& 0.9344	& 0.9344	& 0.0001	& $0.7291\pm0.0856$	& 6	& 40	& 97\\
16	& 0.9356	& 0.9356	& 0.0000	& $0.8016\pm0.0648$	& 6	& 41	& 69\\
17	& 0.9447	& 0.9447	& 0.0002	& $0.6717\pm0.2044$	& 5	& 23	& 70\\
18	& 0.8921	& 0.8922	& 0.0005	& $0.5998\pm0.1614$	& 5	& 28	& 59\\
19	& 0.9021	& 0.9022	& 0.0009	& $0.4796\pm0.2666$	& 5	& 15	& 36\\
20	& 0.8590	& 0.8591	& 0.0003	& $0.6657\pm0.1221$	& 6	& 21	& 21\\
21	& 0.9286	& 0.9286	& 0.0004	& $0.6255\pm0.2003$	& 5	& 17	& 60\\
  \end{tabular}

\spacer
  \caption{S-based inter-annotator agreements and pairwise mean \oneminuswd{}
  and standard deviation
  with the number of coders, boundaries, and mass per chapter}
  \label{table:agreements}
\spacersmall
\spacer
\spacer
\end{table}

\section{Conclusion and Future Work}\label{section:conclusion}
We have proposed a segmentation evaluation metric which solves the key
problems facing segmentation analysis today, including an inability to:
appropriately quantify near misses when evaluating automatic segmenters and
human performance; penalize errors equally (or, with configuration, in
a manner that suits a specific segmentation task); compare an automatic
segmenter directly against human performance; require a ``true''
reference; and handle multiple boundary types. Using \metricacronym{},
task-specific evaluation of automatic and human segmenters can be performed
using multiple human judgements unhindered by the quirks of window-based
metrics.

In current and future work, we will show how S can be used to analyze
hierarchical segmentations, and illustrate how to apply S to linear
segmentations containing multiple boundary types.

\section*{Acknowledgments}
We thank Anna Kazantseva for her invaluable feedback and
corpora, and Stan Szpakowicz, Martin Scaiano, and James Cracknell for their
feedback.

\end{document}